%% file: Paper2.tex
\documentclass[conference]{./IEEEtran}
\usepackage[left=52pt, right=52pt, top=54pt, bottom=54pt]{geometry}

\usepackage{cite}

\ifCLASSINFOpdf
  \usepackage[pdftex]{graphicx}
\else
\fi
\usepackage{amsmath}
\usepackage{array}

\ifCLASSOPTIONcompsoc
 \usepackage[caption=false,font=normalsize,labelfont=sf,textfont=sf]{subfig}
\else
 \usepackage[caption=false,font=footnotesize]{subfig}
\fi
\usepackage{epstopdf}

\usepackage[ampersand]{easylist}
\usepackage{amssymb}
\ListProperties(Hide=100, Hang=true, Progressive=3ex, Style*=-- ,
Style2*=$\bullet$ ,Style3*=$\circ$ ,Style4*=\tiny$\blacksquare$)

\usepackage{tabularx}
\usepackage{dblfloatfix}    

\DeclareMathOperator*{\argmax}{arg\,max}

\usepackage{makecell}
\usepackage{booktabs}

\begin{document}
%

\title{Automated Speed and Lane Change Decision Making using Deep Reinforcement Learning}


%

\author{\IEEEauthorblockN{Carl-Johan Hoel\IEEEauthorrefmark{1}\IEEEauthorrefmark{2},
Krister Wolff\IEEEauthorrefmark{1},
Leo Laine\IEEEauthorrefmark{1}\IEEEauthorrefmark{2}}
\IEEEauthorblockA{
\IEEEauthorrefmark{1}Chalmers University of Technology, 412 96 G\"oteborg, Sweden\\
\IEEEauthorrefmark{2}Volvo Group Trucks Technology, 405 08 G\"oteborg, Sweden\\
 Email: \{carl-johan.hoel, krister.wolff, leo.laine\}@chalmers.se}
}

\maketitle

\begin{abstract}

This paper introduces a method, based on deep reinforcement learning, for automatically generating a general purpose decision making function. A Deep Q-Network agent was trained in a simulated environment to handle speed and lane change decisions for a truck-trailer combination. In a highway driving case, it is shown that the method produced an agent that matched or surpassed the performance of a commonly used reference model. To demonstrate the generality of the method, the exact same algorithm was also tested by training it for an overtaking case on a road with oncoming traffic. Furthermore, a novel way of applying a convolutional neural network to high level input that represents interchangeable objects is also introduced.

\end{abstract}


%
\IEEEpeerreviewmaketitle

\section{Introduction}
\label{sec:introduction}

%


By automating heavy vehicles, there is potential for a significant productivity increase, see e.g. \cite{FAGNANT2015167}. One of the challenges in developing autonomous vehicles is that they need to make decisions in complex environments, ranging from highway driving to less structured areas inside cities.
To predict all possible traffic situations, and code how to handle them, would be a time consuming and error prone work, if at all feasible. Therefore, a method that can learn a suitable behavior from its own experiences would be desirable. Ideally, such a method should be applicable to all possible environments. This paper introduces how a specific machine learning algorithm can be applied to automated driving, here tested on a highway driving case and an overtaking case.

%

Traditionally, rule based gap acceptance models are common to make lane changing decisions, see for example \cite{Gipps} or \cite{Ahmed}.
More recent methods often consider the utility of a potential lane change.
Either the utility of changing to a specific lane is estimated, see \cite{Eggert} or \cite{nilsson}, or the total utility (also called the expected return) over a time horizon is maximized by solving a partially observable Markov decisions process (POMDP), see \cite{Ulbrich} or \cite{sisl}.
Two commonly used models for speed control and to decide when to change lanes are the Intelligent driver model (IDM) \cite{IDM} and the Minimize overall braking induced by lane changes (MOBIL) model \cite{MOBIL}. The combination of these two models was used as a baseline when evaluating the method presented in this paper.

A common problem with most existing methods for autonomous driving is that they target one specific driving case. For example, the ones mentioned above are designed for highway driving, but if a different case is considered, such as driving on a road with oncoming traffic, a completely different method is required.
In an attempt to overcome this issue, we introduced a more general approach in \cite{Paper1}. This method is based on a genetic algorithm, which is used to automatically train a general-purpose driver model that can handle different cases. However, the method still requires some features to be defined manually, in order to adapt its rules and actions to different driving cases.

During the last years, the field of deep learning has made revolutionary progress in many areas, see e.g.~\cite{Schmidhuber2015} or \cite{LeCun2015}. By combining deep neural networks with reinforcement learning, artificial intelligence has evolved in different domains, from playing Atari games \cite{atari}, to continuous control \cite{Lillicrap2015}, reaching a super human performance in the game of Go \cite{Silver2017go} and beating the best chess computers \cite{Silver2017chess}. Deep reinforcement learning has also successfully been used for some special applications in the field of autonomous driving, see e.g. \cite{Shalev2016} and \cite{Sallab:2017:2470-1173:70}.

This paper introduces a method based on a Deep Q-Network (DQN) agent \cite{atari} that, from training in a simulated environment, automatically generates a decision making function. To the extent of the authors' knowledge, this method has not previously been applied to this problem.
The main benefit of the presented method is that it is general, i.e. not limited to a specific driving case. For highway driving, it is shown that it can generate an agent that performs better than the combination of the IDM and MOBIL model. Furthermore, with no tuning, the same method can be applied to a different setting, in this case driving on a road with oncoming traffic. Two important differences compared to our previous approach in \cite{Paper1} is that the method presented in this paper does not need any hand crafted features and that the training is significantly faster. Moreover, this paper introduces a novel way of using a convolutional neural network architecture by applying it to high level sensor data, representing interchangeable objects, which improves and speeds up the learning process.

This paper is organized as follows: The DQN algorithm and how it was implemented is described in Sect.~\ref{sec:method}. Next, Sect.~\ref{sec:simulations} gives an overview of the IDM and the MOBIL model, and describes how the simulations were set up. In Sect.~\ref{sec:results}, the results are presented, followed by a discussion in Sect.~\ref{sec:discussion}. Finally the conclusions are given in Sect.~\ref{sec:conclusion}.


\section{Speed and lane change decision making}
\label{sec:method}

In this paper, the task of deciding when to change lanes and to control the speed of the vehicle under consideration
(henceforth referred to as the ego vehicle) is viewed as a reinforcement learning problem. A Deep Q-Network (DQN) agent \cite{atari} is used to learn the Q-function, which describes how beneficial different actions are in a given state. The state of the surrounding vehicles and the available lanes are known to the agent, and its objective is to choose which action to take, which for example could be to change lanes, brake or accelerate.
The details of the procedure are described in this section.

\subsection{Reinforcement learning}

Reinforcement learning is a branch of machine learning, where an agent acts in an environment and tries to learn a policy, $\pi$, that maximizes a cumulative reward function. 
The policy defines which action, $a$, to take, given a state, $s$. The state of the environment will then change to a new state, $s'$, and return a reward, $r$.
The reinforcement learning problem is often modeled as a Markov Decision Process (MDP), which is defined as the tuple $\langle S, A, T, R, \gamma \rangle$, where $S$ is the set of states, $A$ is the set of actions, $T:S \times A \rightarrow S$ is the state transition probability function, $R:S \times A \times S \rightarrow \mathbb{R}$ is the reward function and $\gamma \in [0,1]$ is a discount factor. An MDP satisfies the Markov property, which means that the probability distribution of the future states depends only on the current state and action, and not on the history of previous states. At every time step, $t$, the goal of the agent is to maximize the future discounted return, defined as
\begin{align}
    R_t = \sum_{k=0}^\infty \gamma^k r_{t+k},
\end{align}
where $r_{t+k}$ is the reward given at step $t+k$.
See \cite{Sutton:1998} for a comprehensive introduction to reinforcement learning and MDPs.

\subsection{Deep Q-Network}
\label{sec:DQN}

In the reinforcement learning algorithm called Q-learning \cite{Watkins1992}, the agent tries to learn the optimal action value function, $Q^*(s,a)$. This function is defined as the maximum expected return when being in a state, $s$, taking some action, $a$, and then following the optimal policy, $\pi^*$. This is described by
\begin{align}
    Q^*(s,a) = \max_\pi \mathbb{E} \left[R_t | s_t = s, a_t = a, \pi\right].
\end{align}
The optimal action value function follows the Bellman equation, see \cite{Watkins1992},
\begin{align}
    Q^*(s,a) = \mathbb{E}\left[r + \gamma \max_{a'} Q^*(s',a')|s,a\right],
\end{align}
which is based on the intuition that if the values of $Q^*(s',a')$ are known, the optimal policy is to select an action, $a'$, that maximizes the expected value of $Q^*(s',a')$.

In the DQN algorithm \cite{atari}, Q-learning is combined with deep learning. A deep neural network with weights $\theta$ is used as a function approximator of the optimal value function, i.e. $Q(s,a;\theta) \approx Q^*(s,a)$. The network is then trained by adjusting its parameters, $\theta_i$, at every iteration, $i$, to minimize the error in the Bellman equation. This is typically done with stochastic gradient descent, where mini-batches with size $M$ of experiences, described by the tuple $e_t = (s_t, a_t, r_t, s_{t+1})$, are drawn from an experience replay memory. The loss function at iteration $i$ is defined as
\begin{align}
    \resizebox{.91\hsize}{!}{$
    L_i(\theta_i) = \mathbb{E}_\mathrm{M} \Big[ (r + \gamma \max_{a'} Q(s',a';\theta_i^-)
    - Q(s,a;\theta_i) )^2 \Big].
    $}
    \label{eq:loss}
\end{align}
Here, $\theta_i^-$ are the network parameters used to calculate the target at iteration $i$. In order to make the learning process more stable, these parameters are held fixed for a number of iterations and then periodically updated with the latest version of the trained parameters, $\theta_i$.
The trade off between exploration and exploitation is handled by following an $\epsilon$-greedy policy. This means that a random action is selected with probability $\epsilon$, and otherwise the action with the highest value is chosen.
For further details on the DQN algorithm, see \cite{atari}.

Q-learning and the DQN algorithm are known to overestimate the action value function under some conditions. A further development is the Double DQN algorithm \cite{Hasselt2016}, which aims to decouple the action selection and action evaluation. This is done by updating Eq.~\ref{eq:loss} to
%
\begin{align}
    L_i(\theta_i) = \mathbb{E}_\mathrm{M} \big[ \big( & r + \gamma Q(s',\argmax_a Q(s',a;\theta_i);\theta_i^-)  \nonumber \\
    & - Q(s,a;\theta_i) \big)^2 \big].
\end{align}

\subsection{Agent implementation}
\label{sec:methodDetails}
The Double DQN algorithm, outlined above, was applied to control a vehicle in two test cases, which are further described in Sect.~\ref{sec:trafficSimulation}. The details of the implementation of the agent are presented below.

\subsubsection{MDP formulation}

Since the intention of other road users cannot be observed, the speed and lane change decision making problem can be modeled as a partially observable Markov decision process (POMDP) \cite{Kaelbling:1998:PAP:1643275.1643301}. To address the partial observability, the POMDP can be approximated by an MDP with a $k$-Markov approximation, where the state consists of the last $k$ observations, $s_t = (o_{t-k+1},o_{t-k+2},\dots,o_t)$ \cite{atari}. However, for the method presented in this paper, it proved sufficient to set $k=1$, i.e. to simply use the last observation.

Two different agents were investigated in this study, called Agent1 and Agent2. They both used the same state input, $s$, defined as a vector with $27$ elements, which contained information on the ego vehicle's speed, existing lanes and states of the $8$ surrounding vehicles. Table \ref{tab:state} shows the configuration of the state (see Sect.~\ref{sec:simulations} for details on how the traffic environment was simulated).

Agent1 only controlled the lane changing decisions, whereas the speed was automatically controlled by the IDM. This gave a direct comparison to the lane change decisions taken by the MOBIL model, in which the speed also was controlled by the IDM (see Sect.~\ref{sec:referenceModel} for details). Agent2 controlled both the lane changing decisions and the speed. Here, the speed was changed by choosing between four different acceleration options: full brake ($-9$~m/s\textsuperscript{2}), medium brake ($-2$~m/s\textsuperscript{2}), maintain speed ($0$~m/s\textsuperscript{2}) and accelerate ($+2$~m/s\textsuperscript{2}). The action spaces of the two agents are given in Table~\ref{tab:actions}. When a decision to change lanes was taken, the intended lane of the lateral control model, described in Sect.~\ref{sec:trafficSimulation}, was changed. Both agents took decisions at an interval of $\Delta t = 1$ s. \looseness=-1

A simple reward function was used. Normally, at every time step, a positive reward was given, based on the distance driven during that interval, $\Delta d$, and normalized as $\Delta d / \Delta d_\mathrm{max}$. Here, $\Delta d_\mathrm{max} = \Delta t v_\mathrm{max}^\mathrm{ego}$, and $v_\mathrm{max}^\mathrm{ego}$ was the maximum possible speed of the ego vehicle.
This part of the reward function implicitly encouraged lane changes to overtake slower vehicles.
However, if a collision occurred, or the ego vehicle drove out of the road (it could choose to change lanes to one that did not exist), a penalizing reward of $-10$ was given and the episode was terminated. If the ego vehicle ended up in a near collision, defined as being one vehicle length ($4.8$ m) from another vehicle, a reward of $-10$ was also given, but the episode was not terminated. Finally, to limit the number of lane changes, a reward of $-1$ was given when a lane changing action was~chosen.
\begin{table}[!bt]
	\renewcommand{\arraystretch}{1.2}
	\caption{State input vector used by the agents. $s_1$, $s_2$ and $s_3$ describe the state of the ego vehicle and the available lanes, whereas $s_{3i+1}$, $s_{3i+2}$ and $s_{3i+3}$, for $i$ = 1,2,...8, represent the state of the surrounding vehicles.}
	\label{tab:state}
	\centering
	\begin{tabular}{@{\hskip3pt}l@{\hskip9pt}l@{\hskip3pt}}
		\toprule
	    $s_1$ & Normalized ego vehicle speed, $v_\mathrm{ego} / v_\mathrm{ego}^\mathrm{max}$\\
		$s_2$ &  $\begin{cases}
             1, & \text{if there is a lane to the left}\\
             0, & \text{otherwise}
             \end{cases}$\\
		$s_3$ & $\begin{cases}
             1, & \text{if there is a lane to the right}\\
             0, & \text{otherwise}
             \end{cases}$\\
		$s_{3i+1}$ & Normalized relative position of vehicle $i$, $\Delta s_i / \Delta s_\mathrm{max}$\\
		$s_{3i+2}$ & Normalized relative speed of vehicle $i$, $\Delta v_i / v_\mathrm{max}$\\
		$s_{3i+3}$ & $\begin{cases}
             \hfill -1, & \text{if vehicle $i$ is two lanes to the right of the ego vehicle}\\
             \hfill -0.5, & \text{if vehicle $i$ is one lane to the right of the ego vehicle}\\
             \hfill 0, & \text{if vehicle $i$ is in the same lane as the ego vehicle}\\
             \hfill 0.5, & \text{if vehicle $i$ is one lane to the left of the ego vehicle}\\
             \hfill 1, & \text{if vehicle $i$ is two lanes to the left of the ego vehicle}
             \end{cases}$\\
		\bottomrule
	\end{tabular}
\end{table}

\begin{table}[!bt]
	\renewcommand{\arraystretch}{1.2}
	\caption{Action spaces of the two agents.}
	\label{tab:actions}
	\centering
	\begin{tabular}{ll}
	\toprule
	    \multicolumn{2}{c}{Agent1}\\
	    \midrule
	    $a_1$ & Stay in current lane\\
	    $a_2$ & Change lanes to the left\\
		$a_3$ & Change lanes to the right\\
		\midrule
		\multicolumn{2}{c}{Agent2}\\
		\midrule
		$a_1$ & Stay in current lane, keep current speed\\
	    $a_2$ & Stay in current lane, accelerate with -2 m/s\textsuperscript{2}\\
		$a_3$ & Stay in current lane, accelerate with -9 m/s\textsuperscript{2}\\
		$a_4$ & Stay in current lane, accelerate with 2 m/s\textsuperscript{2}\\
		$a_5$ & Change lanes to the left, keep current speed\\
		$a_6$ & Change lanes to the right, keep current speed\\
		\bottomrule
	\end{tabular}
\end{table}

\subsubsection{Neural network design}
Two different neural network architectures were investigated in this study. Both had $27$ input neurons, for the state described above. The final output layer had $3$ output neurons for Agent1 and $6$ output neurons for Agent2, where the value of neuron $n_i$ represented the value function when choosing action $a_i$, i.e. $Q(s,a_i)$.

The first architecture was a standard fully connected neural network (FCNN), with two hidden layers. Each layer consisted of $n_\mathrm{hidden}$ neurons, set to $512$, and rectified linear units (ReLUs) were used as activation functions \cite{Nair:2010:RLU:3104322.3104425}. The final output layer used a linear activation function.

The second architecture introduces a new way of applying temporal convolutional neural networks (CNNs). CNNs are inspired by the structure of the visual cortex in animals. By their architecture and weight sharing properties, they create a space and shift invariance, and reduce the number of parameters to be optimized. This has made them successful in the field of computer vision, where they have been applied directly to low level input, consisting of pixel values. For further details on CNNs, see e.g. \cite{LeCun2015}.

\begin{figure}[!tb]
	\centering
		\includegraphics[width=\columnwidth]{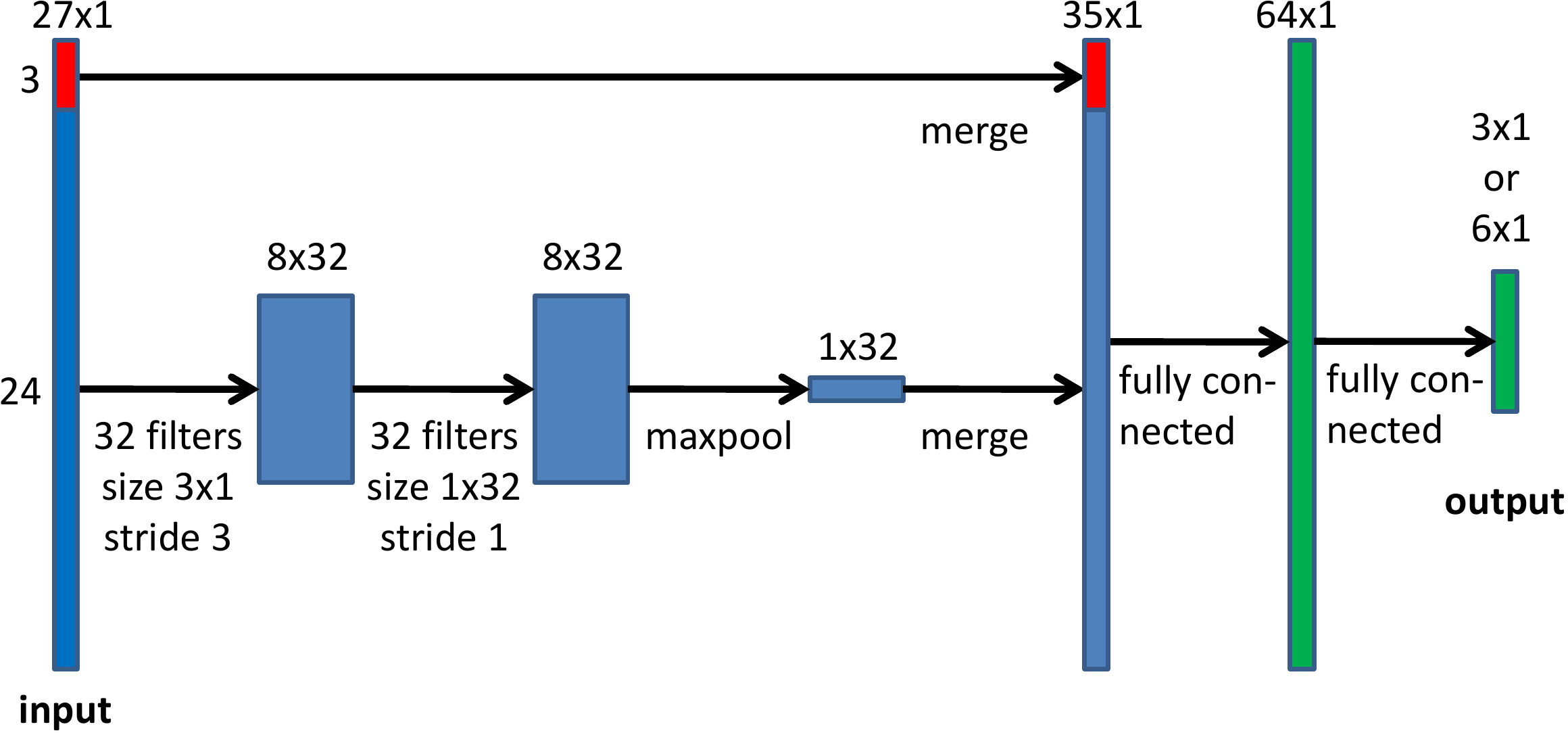}
	\caption{The second network architecture, which used convolutional neural networks and max pooling to create translational invariance between the input from different surrounding vehicles. See the main text for further explanations.}
	\label{fig:networkArchitecture}
\end{figure}

In this study, a CNN architecture was applied to a high level input, which described the state of identical, interchangeable objects, see Fig.~\ref{fig:networkArchitecture}. Two convolutional layers were applied to the part of the state vector that represented the relative position, speed and lane of the surrounding vehicles.
The first layer had $n_\mathrm{conv1}$ filters, set to $32$, with filter size $3$, stride $3$ and ReLU activation functions. This structure created an output of $8\times32$ signals. Since there were $3$ neighbouring input neurons that described the properties of each of the $8$ surrounding vehicles, by setting the filter size and stride to $3$, each row of the output only depended on one vehicle.
The second layer had $n_\mathrm{conv2}$ filters, set to $32$, with filter size $1$, stride $1$ and ReLU activation functions. This further aggregated knowledge about each vehicle in every row of the $8\times32$ output signal.
After the second convolutional layer, a max pooling layer was added.
This structure created a translational invariance of the input that described the relative state of the different vehicles, i.e. the result would be the same if e.g. the input describing vehicle 3 and vehicle 4 switched position in the input vector. This translational invariance, in combination with the reduced number of optimizable parameters, simplified and sped up the training of the network. See Sect.~\ref{sec:discussion} for a further discussion on why a CNN architecture was beneficial in this setting.

The output of the max pooling layer was then concatenated with the rest of the input vector. A fully connected layer with $n_\mathrm{full}$ units, here set to $64$, and ReLu activation functions followed. Finally, the output layer had $3$ or $6$ neurons, both with linear activation functions.

\subsubsection{Training details}

The network was trained by using the Double DQN algorithm, described in Sect.~\ref{sec:DQN}. 
During training, the policy followed an $\epsilon$-greedy behavior, where $\epsilon$ decreased linearly from $\epsilon_{start}$ to $\epsilon_\mathrm{end}$ over $N_\mathrm{\epsilon-end}$ iterations. A discount factor, $\gamma$, was used for future rewards. The target network was updated every $N_\mathrm{update}$ iterations by cloning the online parameters, i.e. setting $\theta_i^- = \theta_i$, at the updating step.
Learning started after $N_\mathrm{start}$ iterations and a replay memory of size $M_\mathrm{replay}$ was used. Mini-batches of training samples with size $M_\mathrm{mini}$ were uniformly drawn from the replay memory and the network was updated using the RMSProp algorithm \cite{RMSProp}, with a learning rate of $\eta$. In order to improve the stability, error clipping was used by limiting the error term $r + \gamma Q(s',\argmax_a Q(s',a;\theta_i);\theta_i^-) - Q(s,a;\theta_i)$ to $[-1,1]$.

The hyperparameters of the training are summarized in Table~\ref{tab:hyperparameters}. Due to the computational complexity, a systematic grid search was not performed. Instead, the hyperparameter values were selected from an informal search, based upon the values given in \cite{atari} and \cite{Hasselt2016}.

The state space, described above, did not provide any information on where in an episode the agent was at a given time step, e.g. if it was in the beginning or close to the end (Sect.~\ref{sec:trafficSimulation} describes how an episode was defined). The reason for this choice was that the goal was to train an agent that performed well in highway driving of infinite length. Therefore, the longitudinal position was irrelevant. However, at the end of a successful episode, the future discounted return, $R_\mathrm{end}$, was $0$. To avoid that the agent learned this, the last experience $e_\mathrm{end}$ was not stored in the experience replay memory. Thereby, the agent was tricked to believe that the episode continued forever.

\begin{table}[!bt]
	\renewcommand{\arraystretch}{1.2}
	\caption{Hyperparameters used to train the DQN agents.}
	\label{tab:hyperparameters}
	\centering
	\begin{tabular}{lc}
		\toprule
		Discount factor, $\gamma$ & $0.99$\\
		Learning start iteration, $N_\mathrm{start}$ & $50{,}000$\\
		Replay memory size, $M_\mathrm{replay}$ & $500{,}000$\\
		Initial exploration constant, $\epsilon_{start}$  & $1$\\
		Final exploration constant, $\epsilon_\mathrm{end}$ & $0.1$\\
		Final exploration iteration, $N_\mathrm{\epsilon{\text -}end}$ & $500{,}000$\\
		Learning rate, $\eta$ & $0.00025$\\
		Mini-batch size, $M_\mathrm{mini}$ & $32$\\
		Target network update frequency, $N_\mathrm{update}$ & $30{,}000$\\
		\bottomrule
	\end{tabular}
\end{table}


\section{Simulation setup}
\label{sec:simulations}

A highway case was used as the main way to test the algorithm outlined above.
To evaluate the performance of the agent, a reference model, consisting of the IDM and MOBIL model, was used.
This section briefly summarizes the reference model, describes how the simulations were set up and how the performance was measured.
Moreover, in order to show the versatility of the proposed method, it was further tested in a secondary overtaking case with oncoming traffic, which is also described here.

\subsection{Reference model}
\label{sec:referenceModel}

The IDM \cite{IDM} is widely used in transportation research to model the longitudinal dynamics of a vehicle. With this model, the speed of the ego vehicle, $v$, varies according to
\begin{align}
\label{eq:IDM}
	\dot{v} = a \bigg( 1 - \left( \frac{v}{v_0} \right)^\delta - \left( \frac{d^*(v,\Delta v)}{d} \right)^2 \bigg),\\
		d^*(v,\Delta v) = d_0 + vT + v\Delta v / (2 \sqrt{ab}).
\end{align}
The vehicle's speed depends on the distance to the vehicle in front, $d$, and the speed difference (approach rate), $\Delta v$. Table~\ref{tab:idmParameters} shows the parameters that are used to tune the model. The values were taken from the original paper \cite{IDM}.

The  MOBIL model \cite{MOBIL} makes decisions on when to change lanes by maximizing the acceleration of the vehicle in consideration and the surrounding vehicles.
For a lane change to be allowed, the induced acceleration of the following car in the new lane, $a_\mathrm{n}$, must fulfill a safety criterion, $a_\mathrm{n} > -b_\mathrm{safe}$. To predict the acceleration of the ego and surrounding vehicles, the IDM model is used. If the safety criterion is met, MOBIL changes lanes if
\begin{align}
	\tilde{a}_\mathrm{e} - a_\mathrm{e} + p \left( (\tilde{a}_\mathrm{n} - a_\mathrm{n}) + (\tilde{a}_\mathrm{o} - a_\mathrm{o}) \right) > a_\mathrm{th},
\end{align}
where  $a_\mathrm{e}$, $a_\mathrm{n}$ and $a_\mathrm{o}$ are the accelerations of the ego vehicle, the trailing vehicle in the target lane, and the trailing vehicle in the current lane, respectively, assuming that the ego vehicle stays in its lane. Furthermore, $\tilde{a}_\mathrm{e}$, $\tilde{a}_\mathrm{n}$ and $\tilde{a}_\mathrm{o}$ are the corresponding accelerations if the lane change is carried out. The politeness factor, $p$, controls how the effect on other vehicles is valued. To perform a lane change, the collective acceleration gain must be higher than a threshold, $\Delta a_\mathrm{th}$.
If there are lanes available both to the left and to the right, the same criterion is applied to both options. If both criteria are fulfilled, the option with the highest acceleration gain is chosen.
The parameter values of the MOBIL model are shown in Table~\ref{tab:idmParameters}. They were taken from the original paper \cite{MOBIL}, except for the politeness factor, here set to $0$.
This setting provided a more fair comparison to the DQN agent, since then neither method considered possible acceleration losses of the surrounding vehicles.

\begin{table}[!bt]
	\renewcommand{\arraystretch}{1.2}
	\caption{IDM and MOBIL model parameters.}
	\label{tab:idmParameters}
	\centering
	\begin{tabular}{lc}
		\toprule
		Minimum gap distance, $d_0$ & $2$ m\\
		Safe time headway, $T$ & $1.6$ s\\
		Maximal acceleration, $a$ & $0.7$ $\mathrm{m/s^2}$\\
		Desired deceleration, $b$ & $1.7$ $\mathrm{m/s^2}$\\
		Acceleration exponent, $\delta$ & $4$\\
		\midrule
		Politeness factor, $p$ & $0$\\
		Changing threshold, $a_\mathrm{th}$ & $0.1$ $\mathrm{m/s^2}$\\
		Maximum safe deceleration, $b_\mathrm{safe}$ & $4$ $\mathrm{m/s^2}$\\
		\bottomrule
	\end{tabular}
\end{table}

\begin{figure*}[!t]
    \centering
            \subfloat[]{\includegraphics[width=1.99\columnwidth]{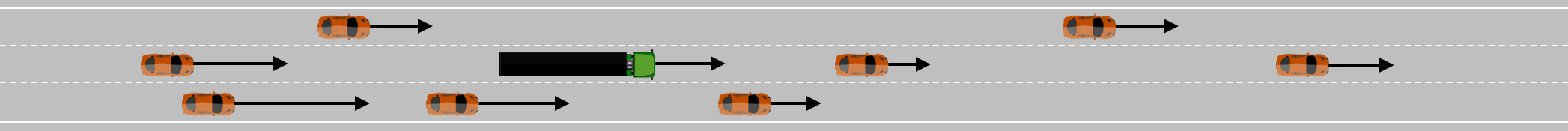}%
            \label{fig:simulationEnvironment}}
        \hfil
            \subfloat[]{\includegraphics[width=1.99\columnwidth]{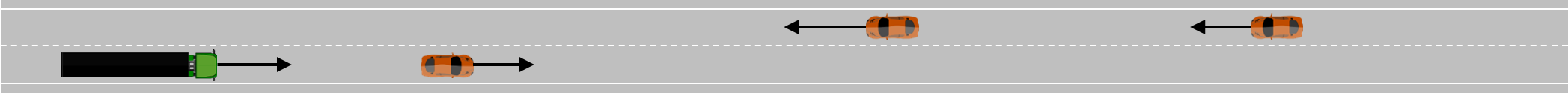}%
            \label{fig:oncoming}}
            \caption{(a) Example of an initial traffic situation for the highway case, which was used as the main way to test the algorithm.
            (b) Example of a traffic situation for a secondary overtaking case with oncoming traffic, showing the situation $10$ seconds from the initial state.
            In both cases, the ego vehicle (truck-trailer combination) is shown in green and black. The arrows represent the velocities of the vehicles.}
        \label{fig:sim}
\end{figure*}

\subsection{Traffic simulation}
\label{sec:trafficSimulation}

\subsubsection{Highway case}
A highway case was used as the main way to test the method presented in this paper. This case was similar to the one used in the previous study \cite{Paper1}. For completeness, it is summarized below.

A three-lane highway was used, where the ego vehicle to be controlled was surrounded by $8$ other vehicles. The ego vehicle consisted of a $16.5$ m long truck-semitrailer combination and the surrounding vehicles were normal $4.8$ m long passenger cars. These surrounding vehicles stayed in their initial lanes and followed the IDM model longitudinally. Overtaking was allowed both on the left and the right side of another vehicle. An example of an initial traffic situation is shown in Fig.~\ref{fig:simulationEnvironment}.

\begin{figure}[!bt]
	\centering
		\includegraphics[width=0.99\columnwidth]{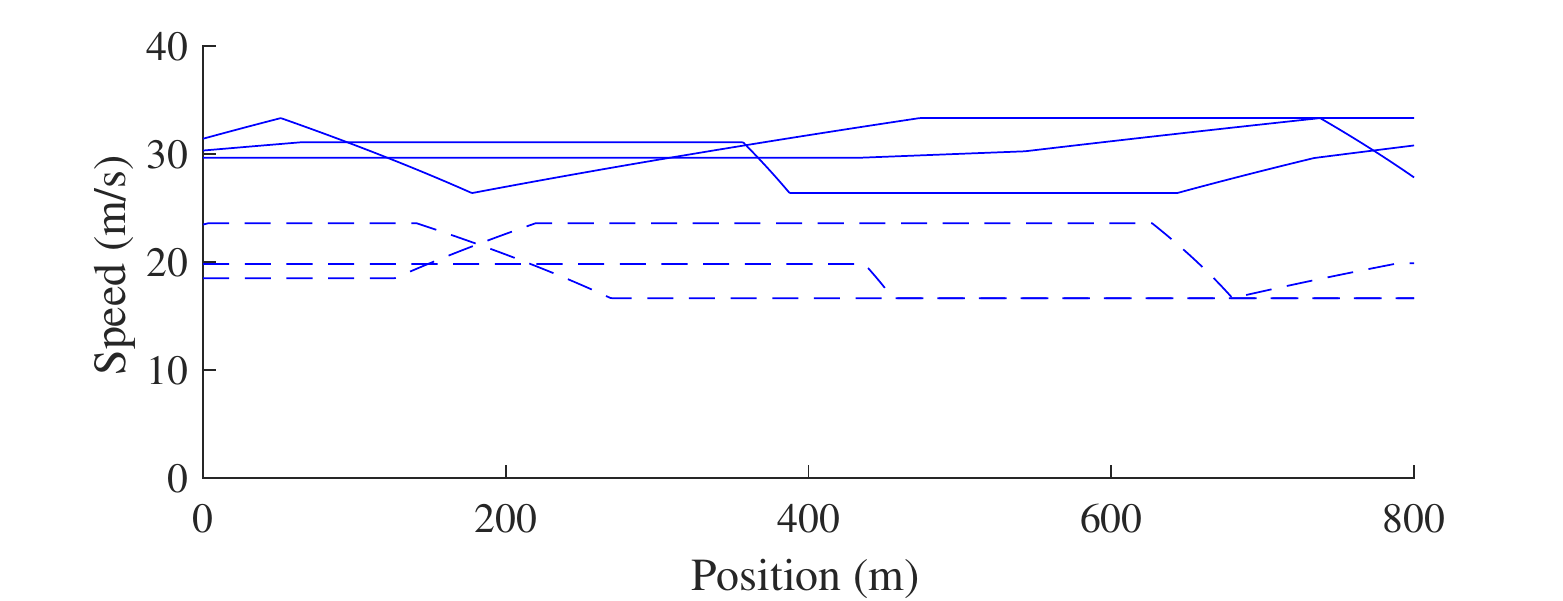}
	\caption{Example of six different randomly generated speed trajectories, defined for different positions along the highway. The solid lines are fast trajectories, applied to vehicles starting behind the ego vehicle, whereas the dashed lines are slow trajectories, applied to vehicles starting in front of the ego vehicle.}
	\label{fig:exampleSpeedTrajectory}
\end{figure}

Although normal highway driving mostly consists of traffic with rather constant speeds and small accelerations, occasionally vehicles brake hard, or even at the maximum of their capability to avoid collisions. Drivers can also decide to suddenly increase their speed rapidly. Therefore, in order for the agent to learn to keep a safe inter-vehicle distance, such quick speed changes need to be included in the training process. The surrounding vehicles in the simulations were assigned different desired speed trajectories. To speed up the training of the agent, these trajectories contained frequent speed changes, which occurred more often than during normal highway driving. Some examples are shown in Fig.~\ref{fig:exampleSpeedTrajectory}.

The ego vehicle initially started in the middle lane, surrounded by $8$ other vehicles. These were randomly positioned in the lanes, within $d_\mathrm{long}$ longitudinally and with a minimum inter-vehicle distance $d_\mathrm{\Delta}$. The initial and maximum ego vehicle speed was $v_\mathrm{init}^\mathrm{ego}$ and $v_\mathrm{max}^\mathrm{ego}$ respectively. Vehicles that were positioned in front of the ego vehicle were assigned slower speed trajectories, in the range $[v^{+}_{\rm{min}}, v^{+}_{\rm{max}}]$, whereas vehicles placed behind the ego vehicle were assigned faster speed trajectories, in the range  $[v^{-}_{\rm{min}}, v^{-}_{\rm{max}}]$.
This created traffic situations where the agent needed to make lane changes to overtake slow vehicles, and at the same time consider faster vehicles approaching from behind.
Episodes where two vehicles were placed too close together with a large speed difference, thus causing an unavoidable collision, were deleted. Each episode was $d_\mathrm{max}$ long.
The values of the mentioned parameters are presented in Table~\ref{tab:trafficSimulationParameters}.
Further details on the setup of the simulations, and how the speed trajectories were generated, are described in \cite{Paper1}.

\begin{table}[!bt]
	\renewcommand{\arraystretch}{1.2}
	\caption{Parameters of the simulated highway case.}
	\label{tab:trafficSimulationParameters}
	\centering
	\begin{tabular}{lc}
		\toprule
		Maximum initial vehicle spread, $d_\mathrm{long}$ & $200$ m\\
		Minimum initial inter-vehicle distance, $d_\mathrm{\Delta}$ & $25$ m\\
		Front vehicle minimum speed, $v^{+}_{\rm{min}}$ & $16.7$ m/s ($60$ km/h)\\
		Front vehicle maximum speed, $v^{+}_{\rm{max}}$ & $23.6$ m/s ($85$ km/h)\\
		Rear vehicle minimum speed, $v^{-}_{\rm{min}}$ & $26.4$ m/s ($95$ km/h)\\
		Rear vehicle maximum speed, $v^{-}_{\rm{max}}$ & $33.3$ m/s ($120$ km/h)\\
		Initial ego vehicle speed, $v_\mathrm{init}^\mathrm{ego}$ & $25$ m/s ($90$ km/h)\\
		Maximum ego vehicle speed, $v_\mathrm{max}^\mathrm{ego}$ & $25$ m/s ($90$ km/h)\\
		Episode length, $d_\mathrm{max}$ & $800$ m\\
		\bottomrule
	\end{tabular}
\end{table}

\subsubsection{Overtaking case}
In order to illustrate the generality of the method presented in this paper, a secondary overtaking case, including two-way traffic, was also tested. Fig.~\ref{fig:oncoming} shows an example of this case.
The ego vehicle started in the right lane, with an initial speed of $v_\mathrm{init}^\mathrm{ego}$, set to $25$ m/s. Another vehicle, which followed a random slow speed profile (defined above), was placed $50$ m in front of the ego vehicle. Two oncoming vehicles, also following slow speed profiles, were placed in the left, oncoming lane, at a random distance between $300$ and $1100$ m in front of the ego vehicle.

\subsubsection{Vehicle motion and lateral control models}
In both the highway and the overtaking case, the motion of the vehicles was simulated by using kinematic models. A lane following two-point visual control model \cite{lateralControl} was used to control the vehicles laterally. As mentioned in Sect.~\ref{sec:methodDetails}, when the agent decided to change lanes, the setpoint of this model was changed to the new desired lane. The same procedure was used if the MOBIL model decided to change lanes. With this control model, a lane change normally took 2 to 3 s, depending on the longitudinal speed. See \cite{Paper1} for further details on the vehicle motion and lateral control models.

\subsection{Performance index}
\label{sec:scoreFunction}

In order to evaluate how the DQN agent performed compared to the reference driver model (presented in Sect.~\ref{sec:referenceModel}) in a specific episode of the highway case, a performance index, $\tilde{p}$, was defined as
\begin{align}
    \tilde{p} = (d/d_\mathrm{max}) (\bar{v}/\bar{v}_\mathrm{ref}).
\end{align}
Here, $d$ is the distance driven by the ego vehicle (limited by a collision or the episode length), $d_\mathrm{max}$ is the episode length, $\bar{v}$ is the average speed of the ego vehicle and $\bar{v}_\mathrm{ref}$ is the average speed when the reference model controlled the ego vehicle through the episode. With this definition, the distance driven by the ego vehicle was the dominant limiting factor when a collision occurred. However, if the agent managed to complete the episode without collisions, the average speed determined the performance index. A value larger than 1 means that the agent performed better than the reference model.

For the overtaking case, the reference model described above cannot be used. Instead, the performance index was simply defines as $\tilde{p}_\mathrm{o} = (d/d_\mathrm{max}) (\bar{v}/\bar{v}_\mathrm{refIDM})$. Here, $\bar{v}_\mathrm{refIDM}$ was the mean speed of the ego vehicle when it was controlled by the IDM through the same episode, i.e. when it did not overtake the preceding vehicle.


\section{Results}
\label{sec:results}


This section focuses on the results that were obtained for the highway case, described in Sect.~\ref{sec:trafficSimulation}, which was the main way of testing the presented method. It also briefly explains and discusses some characteristics of the results, whereas a more general discussion follows in Sect.~\ref{sec:discussion}. The results regarding the overtaking case are collected in Sect.~\ref{sec:resultsOvertaking}.

As described in Sect.~\ref{sec:method}, two agents with different action spaces were investigated. Agent1 only decided when to change lanes, whereas Agent2 decided both the speed and when to change lanes. Furthermore, two different neural network architectures were used. In summary, the four variants were Agent1\textsubscript{FCNN}, Agent1\textsubscript{CNN}, Agent2\textsubscript{FCNN} and Agent2\textsubscript{CNN}.

Five different runs were carried out for the four agent variants, where each run had different random seeds for the DQN and the traffic simulation. The networks were trained for $2$ million iterations ($3$ million for Agent2\textsubscript{FCNN}), and at every $50{,}000$ iterations, they were evaluated over $1{,}000$ random episodes. Note that these evaluation episodes were randomly generated, and not presented to the agents during training. During the evaluation runs, the performance index described in Sect.~\ref{sec:scoreFunction} was used to compare the agents' and the reference model's behaviour.
The results are shown in Fig.~\ref{fig:completion}, which presents the average proportion, $\hat{p}$, of successfully completed, i.e. collision free, evaluation episodes of the four agent variants, and in Fig.~\ref{fig:score}, which shows their average performance index, $\tilde{p}$. The final performance of the fully trained agents is summarized in Table~\ref{tab:results}.

\begin{table}[!bt]
	\renewcommand{\arraystretch}{1.2}
	\caption{Summary of the results of the different agents for the highway case and the overtaking case.}
	\label{tab:results}
	\centering
	\begin{tabular}{@{\hskip3pt}l@{\hskip8pt}c@{\hskip9pt}c@{\hskip9pt}c@{\hskip9pt}c@{\hskip3pt}}
	\toprule
	    & \multicolumn{2}{c}{Highway case} & \multicolumn{2}{c}{Overtaking case}\\
	    \midrule
	    & \makecell{Collision free\\ episodes} & \makecell{Performance\\ index, $\tilde{p}$} & \makecell{Collision free\\ episodes} & \makecell{Performance\\ index, $\tilde{p}_\mathrm{o}$}\\
	    \midrule
	    Agent1\textsubscript{CNN} & $100\%$ & $1.01$ & $100\%$ & $1.06$\\
	    Agent2\textsubscript{CNN} & $100\%$ & $1.10$ & $100\%$ & $1.11$\\
	    Agent1\textsubscript{FCNN} & $98\%$ & $0.98$ & - & -\\
	    Agent2\textsubscript{FCNN} & $86\%$ & $0.96$ & - & -\\
		\bottomrule
	\end{tabular}
\end{table}

\subsection{Agents using a CNN}

\begin{figure}[!tb]
		\includegraphics[width=\columnwidth]{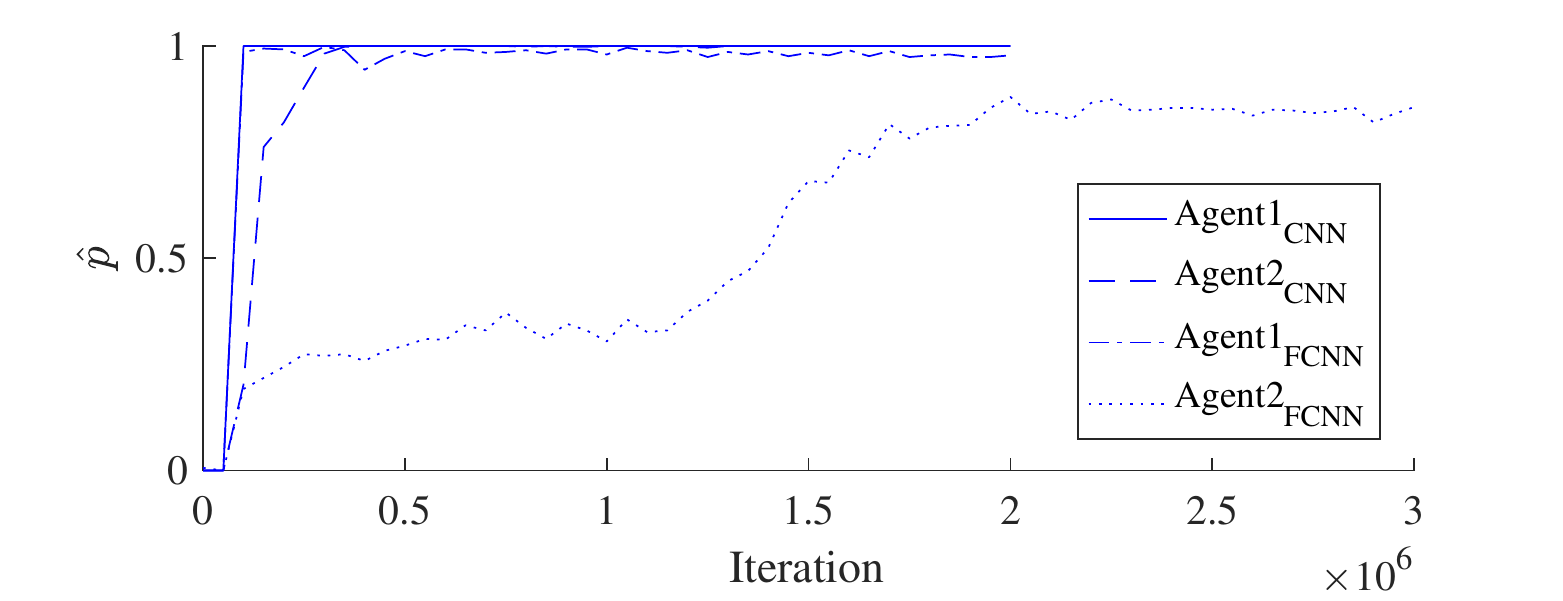}
	\caption{Proportion of episodes solved without collisions by the different agents during training.}
	\label{fig:completion}
\end{figure}

\begin{figure}[!tb]
		\includegraphics[width=\columnwidth]{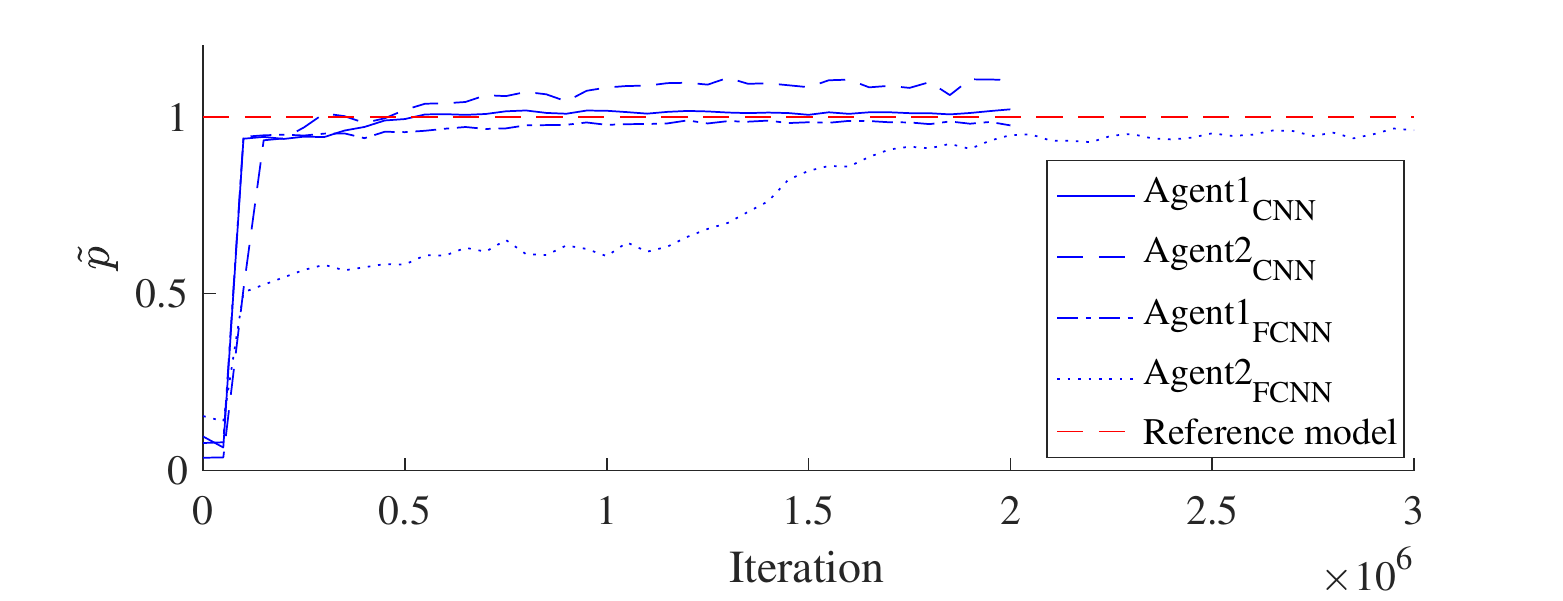}
	\caption{Performance index of the different agents during training.}
	\label{fig:score}
\end{figure}

In Fig.~\ref{fig:completion}, it can be seen that Agent1\textsubscript{CNN} solved all the episodes already after $100{,}000$ iterations, which is the first evaluation after that the training started at $50{,}000$ iterations. At this point it had learned to always stay in its lane, in order to avoid collisions. Since it often got blocked by slower vehicles, its average performance index was therefore lower than $1$ at this point, see Fig.~\ref{fig:score}. However, after around $600{,}000$ iterations, Agent1\textsubscript{CNN} had learned to carry out lane changes when necessary, and performed similar to the reference model.

Fig.~\ref{fig:completion} shows that Agent2\textsubscript{CNN} quickly figured out how to change lanes and increase its speed to solve most of the episodes. Its performance index was on par with the reference model (reached 1) early on during the training, at around $250{,}000$ iterations, see Fig.~\ref{fig:score}. Then, at $400{,}000$ iterations, it solved all the evaluation episodes without collisions. With more training, there were still no collisions, but the performance index increased and stabilized at $1.1$.

\begin{figure}[!tb]
		\includegraphics[width=0.49\columnwidth]{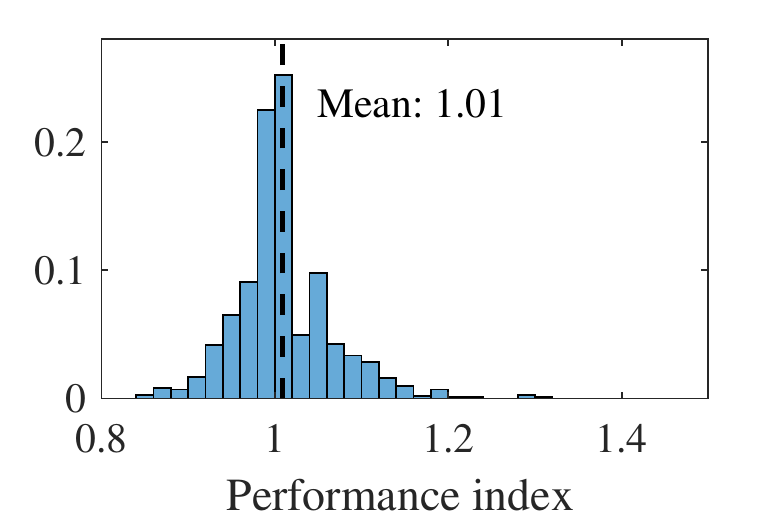}
		\includegraphics[width=0.49\columnwidth]{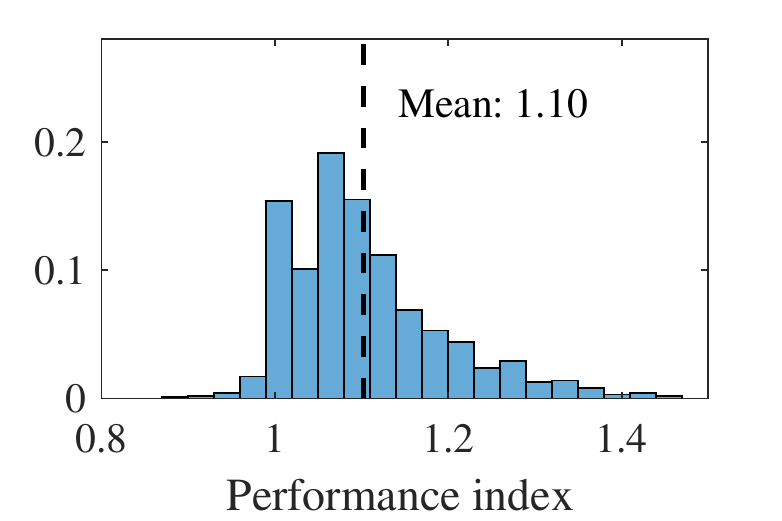}
	\caption{Histogram of the performance index at the end of the training for Agent1\textsubscript{CNN} (left) and Agent2\textsubscript{CNN} (right).}
	\label{fig:histogram}
\end{figure}

Fig.~\ref{fig:histogram} shows a histogram of the performance index for $1{,}000$ evaluation episodes, which were run by the final trained version of Agent1\textsubscript{CNN} and Agent2\textsubscript{CNN}. Since all the episodes were completed without collisions, the performance index was simply the speed ratio $\bar{v} / \bar{v}_\mathrm{ref}$. In the figure, it can be seen that most often there was a small difference between the average speed of the agents and the reference model. There were also some outliers, which were both faster and slower than the reference model. The explanation for these is that the episodes were randomly generated, which meant that even a reasonable action could get the ego vehicle into a situation where it got locked in and could not overtake the surrounding vehicles. Therefore, a small difference in behaviour could lead to such situations for both the trained agents and the reference model, which explains the outliers. Furthermore, the peak at index $1$ for Agent2\textsubscript{CNN} is explained by that there were some episodes when the lane in front of the ego vehicle was free from the start. Then both the reference model and the agents drove at the maximum speed through the whole episode.

\begin{figure}[!tb]
		\includegraphics[width=\columnwidth]{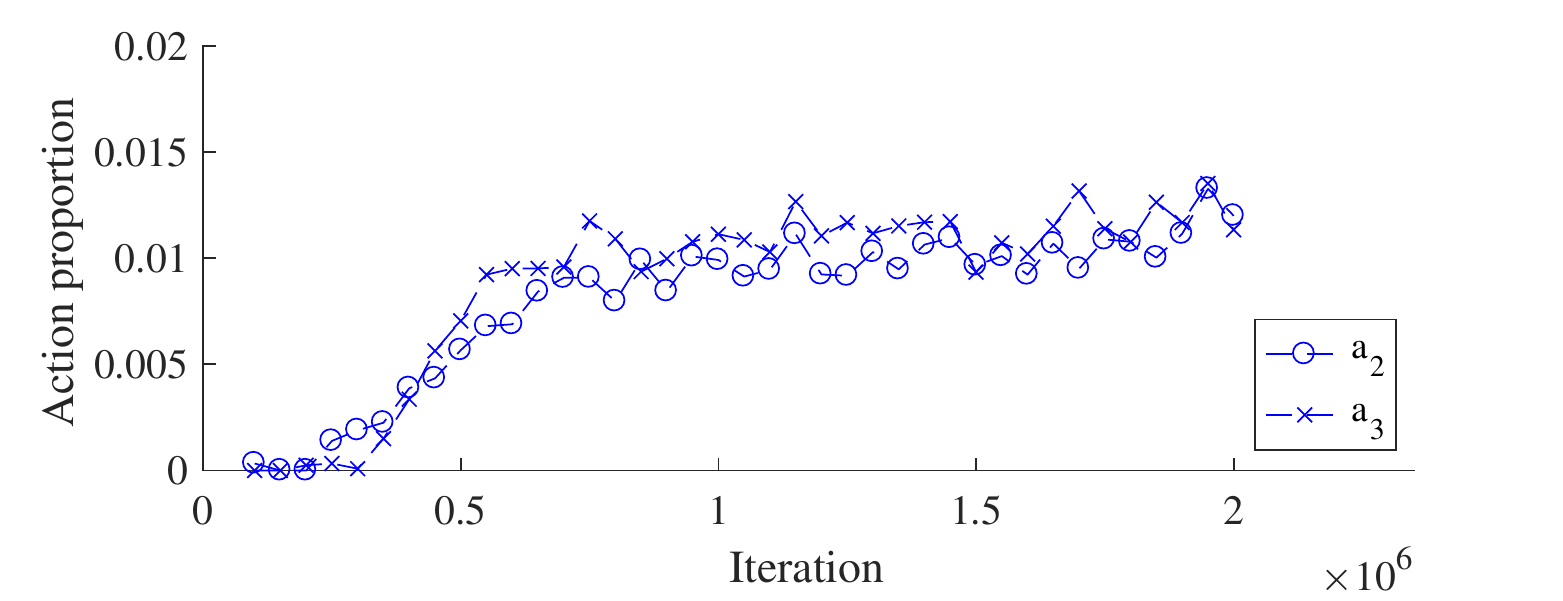}
		\includegraphics[width=\columnwidth]{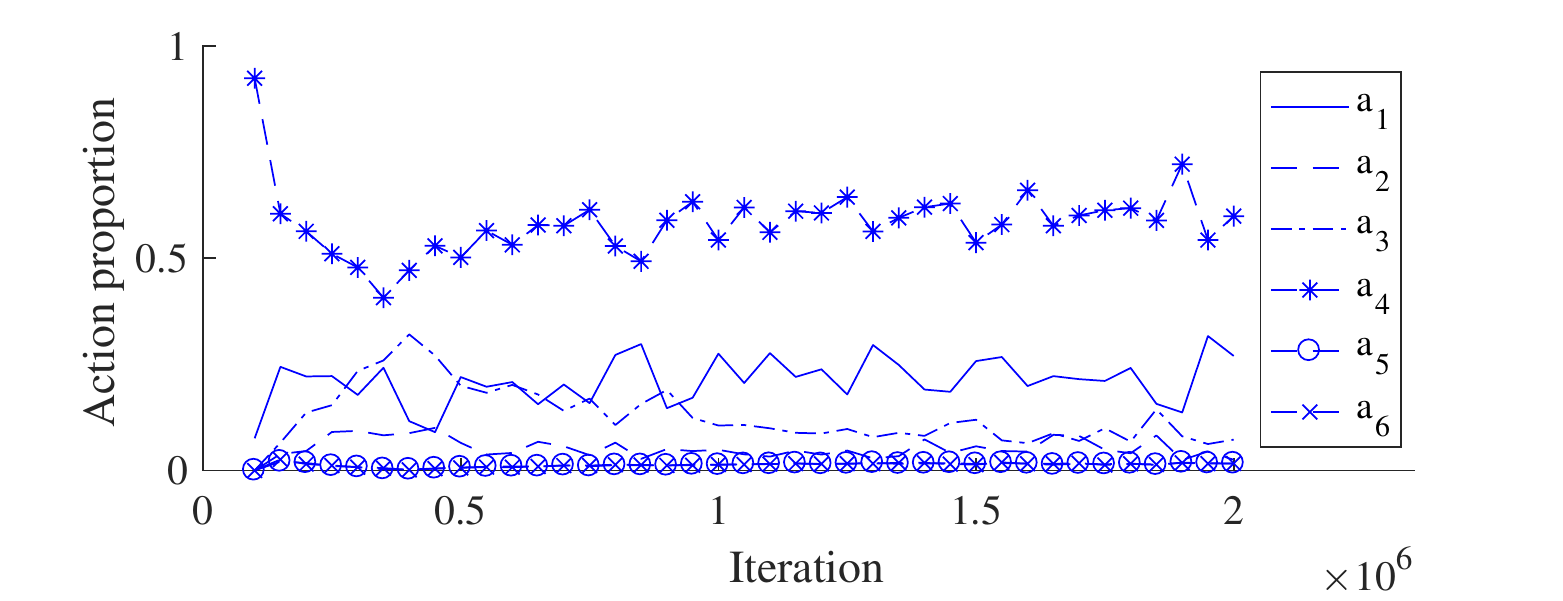}
	\caption{Top: proportion of actions chosen by Agent1\textsubscript{CNN} during training. Due to the scale difference, $a_1$, i.e. stay in the current lane, is here left out. Bottom: proportion of actions chosen by Agent2\textsubscript{CNN} during training. Both plots start at $100{,}000$ iterations, since that is the first evaluation point after that training started at $50{,}000$ iterations.}
	\label{fig:actions}
\end{figure}

To further illustrate the properties of the agents, and how they developed during training, the percentage of chosen actions is shown in Fig.~\ref{fig:actions}. For Agent1\textsubscript{CNN}, it can be seen that it quickly figured out that changing lanes can lead to collisions, and therefore it chose to stay in its lane almost $100\%$ of the time in the beginning. This explains why it completed all the episodes already from the first evaluation point after its training started. However, as training proceeded, it figured out when it safely could change lanes, and thereby perform better. At the end of its training, it chose to change lanes around $1\%$ of the time.
Agent2\textsubscript{CNN} first learned a short sighted strategy, where it accelerated most of the time to obtain a high immediate reward. This naturally led to many rear end collisions. However, when its training proceeded, it learned to control its speed by braking or idling, and to change lanes when necessary.
Reassuringly, both agents learned to change lanes to the left and right equally often.

\subsection{Agents using a FCNN}

Both Agent1\textsubscript{FCNN} and Agent2\textsubscript{FCNN} failed to complete all the evaluation episodes without collisions, see Fig.~\ref{fig:completion} and Table~\ref{tab:results}. Naturally, Agent1\textsubscript{FCNN} solved a significantly higher fraction of the episodes and performed better than Agent2\textsubscript{FCNN}, since it only needed to decide when to change lanes, and not control the speed. In the beginning, it learned to always stay in its lane, and thereby solved all episodes without collisions, but reached a lower performance index than the reference model, see Fig.~\ref{fig:score}. With more training, it started to change lanes and performed reasonably well, but sometimes caused collisions. Agent2\textsubscript{FCNN} performed significantly worse and collided in $14\%$ of the episodes by the end of its training. A longer training run was carried out for Agent1\textsubscript{FCNN} and Agent2\textsubscript{FCNN}, but after $20$ million iterations, the results were the same.

\subsection{Overtaking case}
\label{sec:resultsOvertaking}

\begin{figure}[!tb]
		\includegraphics[width=0.97\columnwidth]{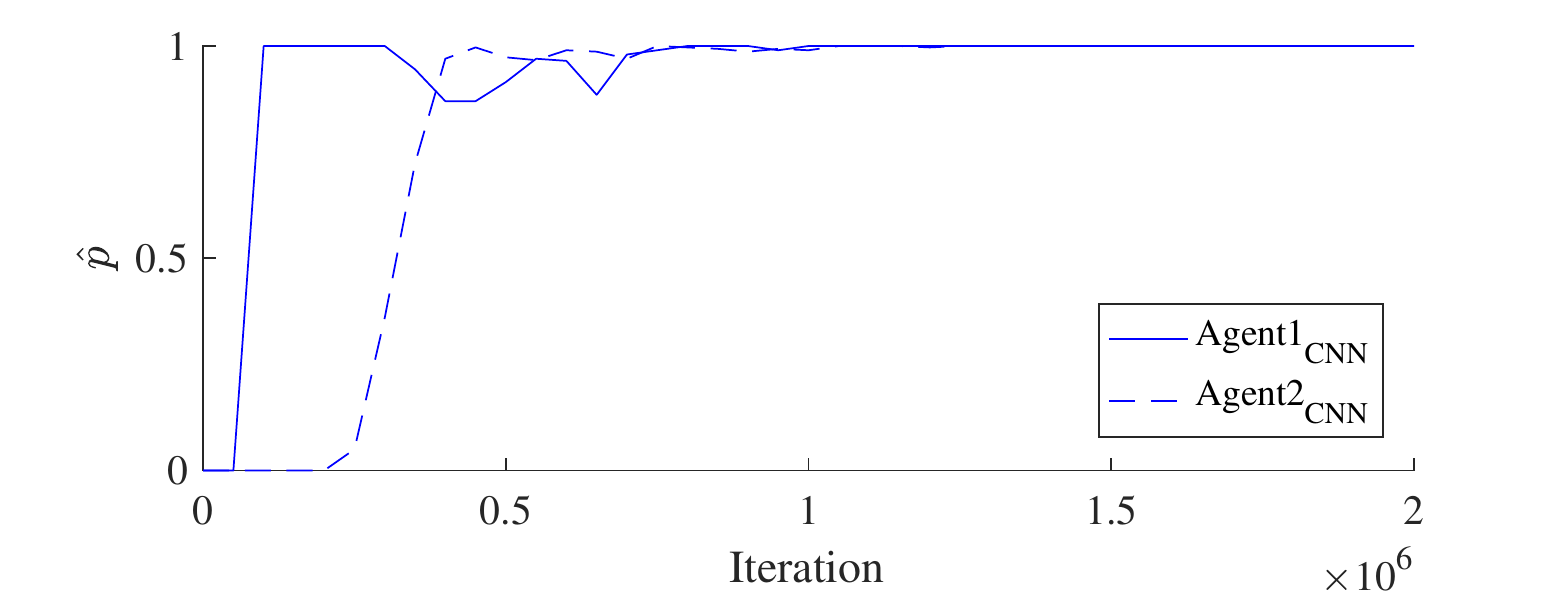}
	\caption{Proportion of overtaking episodes solved without collisions by the different agents during training.}
	\label{fig:completionOvertaking}
\end{figure}

\begin{figure}[!tb]
		\includegraphics[width=0.97\columnwidth]{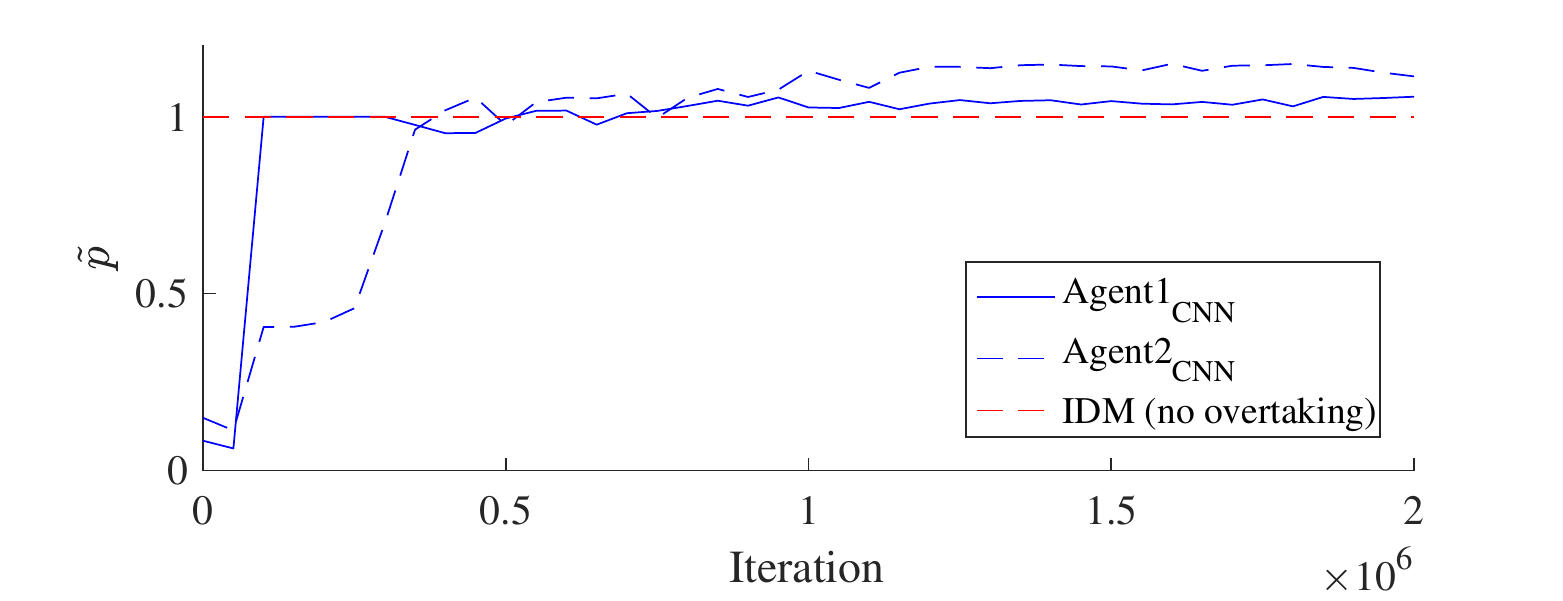}
	\caption{Performance index of the different agents during training on the overtaking case.}
	\label{fig:scoreOvertaking}
\end{figure}

In order to demonstrate the generality of the method presented in this paper, the same algorithm was applied to an overtaking situation, described in Sect.~\ref{sec:trafficSimulation}.
Fig.~\ref{fig:completionOvertaking}, Fig.~\ref{fig:scoreOvertaking} and Table~\ref{tab:results} show the proportion of successfully completed evaluation episodes, $\hat{p}$, and the modified performance index, $\tilde{p}_\mathrm{o}$, of Agent1\textsubscript{CNN} and Agent2\textsubscript{CNN}. By the end of the training, both agents solved all episodes without collisions. Furthermore, in all the episodes, the ego vehicle overtook the slower vehicle, resulting in performance indexes above $1$.

\section{Discussion}
\label{sec:discussion}


In Table~\ref{tab:results}, it can be seen that both Agent1 and Agent2 with the convolutional neural network architecture solved all the episodes without collisions. The performance of Agent1\textsubscript{CNN} was on par with the reference model.
Since they both used the IDM to control the speed, this result indicates that the trained agent and the MOBIL model took lane changing decisions with similar quality. However, when adding the possibility for the agent to also control its speed, as in Agent2\textsubscript{CNN}, the trained agent had the freedom to find better strategies and could therefore outperform the reference model. This result illustrates that for a better performance, lateral and longitudinal decisions should not be completely separated. 

As expected, using a CNN architecture resulted in a significantly better performance than a FCNN architecture, see e.g. Table~\ref{tab:results}. The reason for this is, as mentioned in Sect.~\ref{sec:methodDetails}, that the CNN architecture creates a translational invariance of the input that describes the relative state of the different vehicles. This is reasonable, since it is desirable that the agent reacts the same way to other vehicles' behaviour, independently of where they are positioned in the input vector. Furthermore, since CNNs share weights, the complexity of the network is reduced, which in itself speeds up the learning process.
This way of using CNNs can be compared to how they previously were introduced and applied to low level input, often on pixels in an image, where they provide a spatial invariance when identifying features, see e.g. \cite{Lecun1998}. The results of this paper show that it can also be beneficial to apply CNNs to high level input of interchangeable objects, such as the state description shown in Sect.~\ref{sec:methodDetails}.

As mentioned in Sect.~\ref{sec:methodDetails}, a simple reward function was used. Naturally, the choice of reward function strongly affects the resulting behaviour. For example, when no penalty was given for a lane change, the agent found solutions where it constantly demanded lane changes in opposite directions, which made the vehicle drive in between two lanes. In this study, a simple reward function worked well, but for other cases a more careful design may be required. One way to determine a reward function that mimics human preferences is to use inverse reinforcement learning \cite{IRL}.

In a previous paper, \cite{Paper1}, we presented a different method, based on a genetic algorithm, that automatically can generate a driving model for similar cases as described here. That method is also general and it was shown that it is applicable to different cases, but it requires some hand crafted features when designing the structure of its rules. However, the method presented in this paper requires no such hand crafted features, and instead uses the measured state, described in Table~\ref{tab:state}, directly as input. Furthermore, the method in \cite{Paper1} achieved a similar performance when it comes to safety and average speed, but the number of necessary training episodes was between one and two orders of magnitude higher than for the method that was investigated in this study. Therefore, the new method is clearly advantageous compared to the previous one.

An important remark is that when training an agent by using the method presented in this paper, the agent will only be able to solve the type of situations that it is exposed to in the simulations. It is therefore important that the design of the simulated traffic environment covers the intended case. Furthermore, when using machine learning to produce a decision making function, it is hard to guarantee functional safety. Therefore, it is common to use an underlying safety layer, which verifies the safety of a planned trajectory before it is executed by the vehicle control system, see e.g. \cite{Underwood2016}.


\section{Conclusion and future work}
\label{sec:conclusion}

The main results of this paper show that a Deep Q-Network agent can be trained to make decisions in autonomous driving, without the need of any hand crafted features. In a highway case, the DQN agents performed on par with, or better than, a reference model based on the IDM and MOBIL model. Furthermore, the generality of the method was demonstrated by applying it to a case with oncoming traffic. In both cases, the trained agents handled all episodes without collisions. Another important conclusion is that, for the presented method, applying a CNN to high level input that represents interchangeable objects can both speed up the learning process and increase the performance of the trained agent.

Topics for future work include to further analyze the generality of this method by applying it to other cases, such as crossings and roundabouts, and to systematically investigate the impact of different parameters and network architectures.
Moreover, it would be interesting to apply prioritized experience replay \cite{Schaul2015PrioritizedER}, which is a method where important experiences are repeated more frequently during the training process. This could potentially improve and speed up the learning process.

\section*{Acknowledgment}
This work was partially supported by the Wallenberg Artificial Intelligence, Autonomous Systems and Software Program (WASP), funded by Knut and Alice Wallenberg Foundation, and partially by Vinnova FFI.




\bibliographystyle{IEEEtran}

\linespread{0.965}
\input{Paper2.bbl}

%
%
%

\end{document}